# A machine learning approach for Premature Coronary Artery Disease Diagnosis according to Different Ethnicities in Iran

Mohamad Roshanzamir[1,c], Roohallah Alizadehsani[2,3,4,c], Ehsan Zarepur[5], Noushin Mohammadifard[6], Fatemeh Nouri[5], Mahdi Roshanzamir[7], Alireza Khosravi[8,9], Fereidoon Nouhi[9,10], Nizal Sarrafzadegan[5,11*]

[1]Department of Computer Engineering, Faculty of Engineering, Fasa University, Fasa, 74617-81189, Iran

[2]Biostatistics Unit, Faculty of Health, Deakin University, Geelong, Australia

[3]The Institute for Mental and Physical Health and Clinical Translation (IMPACT), School of Medicine, Deakin University, Geelong, Australia

[4] Faculty of Health, School of Clinical Sciences, Queensland University of Technology, Brisbane, Australia

[5] Isfahan Cardiovascular Research Center, Cardiovascular Research Institute, Isfahan University of Medical Sciences, Isfahan, Iran

[6] Interventional Cardiology Research Center, Cardiovascular Research Institute, Isfahan University of Medical Sciences, Isfahan, Iran

[7]Department of Electrical and Computer Engineering, University of Tabriz, Tabriz, Iran

[8] Hypertension Research Center, Cardiovascular Research Institute, Isfahan University of Medical Sciences, Isfahan, Iran

[9]Iranian Network of Cardiovascular Research, Iran

[10] Rajaie Cardiovascular Medical and Research Center, Iran University of Medical Sciences, Tehran, Iran

[11]Faculty of Medicine, School of Population and Public Health, University of British Columbia, Vancouver, Canada

c: Same contribution

* **Corresponding author:** Nizal Sarrafzadegan, MD

**Corresponding Address:** Isfahan Cardiovascular Research Center, Cardiovascular Research Institute, Isfahan University of Medical Sciences, Isfahan, Iran

**Email:** nsarrafzadegan@gmail.com

## Abstract

Background

Premature coronary artery disease (PCAD) refers to the early onset of the disease, usually before the age of 55 for men and 65 for women. Coronary Artery Disease (CAD) develops when coronary arteries, the major blood vessels supplying the heart with blood, oxygen, and nutrients, become clogged or diseased. This is often due to many risk factors, including lifestyle and cardiometabolic ones, but few studies were done on ethnicity as one of these risk factors, especially in PCAD.

Method

In this study, we tested the rank of ethnicity among the major risk factors of PCAD, including age, gender, body mass index (BMI), visceral obesity presented as waist circumference (WC), diabetes mellitus (DM), high blood pressure (HBP), high low-density lipoprotein cholesterol (LDL-C), and smoking in a large national sample of patients with PCAD from different ethnicities. All patients who met the age criteria underwent coronary angiography to confirm CAD diagnosis. The weight of ethnicity was compared to the other eight features using feature weighting algorithms in PCAD diagnosis. In addition, we conducted an experiment where we ran predictive models (classification algorithms) to predict PCAD. We compared the performance of these models under two conditions: we trained the classification algorithms, including or excluding ethnicity.

Result

This study analyzed various factors to determine their predictive power influencing PCAD prediction. Among these factors, gender and age were the most significant predictors, with ethnicity being the third most important. The results also showed that if ethnicity is used as one of the input risk factors for classification algorithms, it can improve their efficiency.

Conclusion

Our results show that ethnicity ranks as an influential factor in predicting PCAD. Therefore, it needs to be addressed in the PCAD diagnostic and preventive measures.

**Keywords:** PCAD, ethnicity, machine learning, feature weighting, feature selection, classification algorithms.

# 1. Introduction

Coronary artery disease (CAD) is a condition that affects the arteries supplying blood to the heart. It is primarily caused by atherosclerosis, a process in which plaque, made up of fats, cholesterol, and other substances, builds up inside the coronary arteries. This build-up narrows the arteries over time, reducing blood flow to the heart muscle and leading to symptoms like chest pain (angina) or myocardial infarction. PCAD refers to premature coronary artery disease that occurs at ages younger than 55 years in men and 65 years in women, respectively. Recently, CAD prediction and diagnosis have been studied extensively using ML methods because traditional statistical methods have limited capacity and some weaknesses in analyzing large datasets (1-3). Therefore, machine learning approaches have been used as a subfield of artificial intelligence to extract valuable patterns and information from raw data. Using these approaches makes it possible to gain suitable knowledge without human input and use it for different purposes. Recent research has shown that machine learning methods can be used for early diagnosis, risk stratification, clinical trial recruitment, and various disease classifications.

Meanwhile, race and ethnicity are related to population-specific and genetic variations and can be important in some diseases. Socioeconomic status and access to healthcare are strongly associated with race and ethnicity (4). For example, cardiac catheterization for CAD diagnosis is more common among African American patients than White patients (5). The existence of various ethnicities has an effect on the development of diseases among different racial and ethnic groups (6). Therefore, it is necessary to know the role of ethnicity in various diseases.

Machine learning algorithms have been used to predict heart failure considering race and ethnicity and to determine how feature importance differs across different race and ethnicity groups (7). Suinesiaputra et al. (8) have done a multi-ethnic study of atherosclerosis using deep learning analysis. They tried to use the ability of machine learning algorithms to enable the extraction of suitable information from large-scale legacy datasets. They trained VGGNet convolutional neural networks to detect landmarks using a transfer learning sequence between two-chamber, four-chamber, and short-axis Magnetic Resonance Imaging (MRI) views. Another multi-ethnic study of atherosclerosis has been done to predict cardiovascular events using machine learning approaches (9). They used the random survival forests technique to identify cardiovascular risk factors.

Another study that considered ethnicity and ML methods was performed to detect how low left ventricular(LV) ejection fraction using the 12-lead electrocardiogram (ECG) varies by race and ethnicity by using a deep learning algorithm (10). The researchers determined whether the algorithm's performance is determined by the derivation population or by racial variation in the ECG. In fact, in this study, they aimed to determine the impact of race on the Convolutional Neural Network (CNN) performance to detect the presence of LV systolic dysfunction from an ECG. In addition, a multi-ethnic study of atherosclerosis used deep neural survival networks for cardiovascular risk prediction (11). Both machine learning and deep learning models were evaluated in this study.

Bubdy et al. (12) have done a multi-ethnic study of atherosclerosis and evaluated the risk prediction models of atrial fibrillation. They evaluated improvements in predicting 5-year atrial fibrillation risk when adding novel candidate variables identified by machine learning algorithms. These variables are age, race/ethnicity, height, weight, systolic and diastolic blood pressure, current smoking, use of antihypertensive medication, diabetes, and NT-proBNP. However, there is an important shortage in this study. They did not rank which variables have the most impact, which is quite understandable for clinicians. To predict 10-year risk of heart failure, Segar et al. developed and validated machine learning–based race-specific models. In their research, race-specific ML models for heart failure risk prediction were developed in the Jackson Heart Study (JHS) cohort (for black race–specific model) and white adults from ARIC (for white race–specific model). They showed that race-specific and machine learning-based heart failure risk models that integrate clinical, laboratory, and biomarker data demonstrated superior performance compared with traditional heart failure risk and non-race-specific machine learning models (13).

For risk prediction across multi-ethnic patients, machine learning models such as logistic regression with L2 penalty and L1 lasso penalty, random forest, gradient boosting machine (GBM), and extreme gradient boosting are used by Ward et al. (14). In their research, an electronic health record (EHR) database from Northern California has been used. They tried to determine 5-year ASCVD risk prediction in Asian and Hispanic subgroups by training different machine learning models, including logistic regression with $L_2$ penalty and $L_1$ lasso penalty, random forest, gradient boosting machine (GBM), and extreme gradient boosting.

In general, it can be stated that most machine learning-based systems are only used for limited race or ethnicity types and also for unique heart disease endpoints. In addition, they commonly investigated race types, not ethnicity, while it is better to develop more generalized machine learning-based systems that use more ethnicity groups. In this research, we investigated eight number of ethnicities in Iran.

Different machine learning-based systems are used for various ethnicities and endpoints in a study that was done to evaluate the performance of an ML risk calculator in predicting cardiovascular disease (CVD) risk compared to the traditional American College of Cardiology/American Heart Association ACC/AHA CVD risk calculator. The researchers aimed to address the limitations of the ACC/AHA guidelines, which may underestimate risk in high-risk individuals and overestimate it in low-risk populations, potentially leading to missed opportunities for preventive therapy. By utilizing a robust dataset from the Multi-Ethnic Study of Atherosclerosis and validating the ML model with an external cohort (FLEMENGHO), the study sought to demonstrate that the ML approach could provide more accurate risk assessments and recommend less drug therapy while missing fewer CVD events (15). Another example is (9). In this article, there are six endpoints, namely heart failure, atrial fibrillation, stroke, all CVD, CAD, and all-cause death. Nine machine learning-based systems are used for these endpoints and different ethnicities. The study's primary goal was to compare machine learning approaches, specifically random survival forests (RSFs), with traditional methods like the Cox Proportional Hazard model (Cox-PHM) and established risk scores for predicting cardiovascular events.

Additionally, the study aimed to identify significant predictors for six cardiovascular clinical outcomes within a large epidemiological cohort, the Multi-Ethnic Study of Atherosclerosis. This research sought to enhance the understanding of cardiovascular risk prediction by utilizing advanced machine learning techniques alongside deep phenotyping, which involves comprehensive evaluations of various disease aspects, to improve accuracy and predictive power for outcomes such as all-cause death, stroke, and heart failure.

Compared to the previously investigated studies, the novelty of our study includes the following:
- Investigating the importance of ethnicity for PCAD prediction.
- Determining the effect of ethnicity in the performance of classification algorithms.

## 2. Dataset

The dataset used in this study was collected between 2019 and 2022 in different regions in Iran. The Medical Ethics Committee in Isfahan, Iran, granted ethical approval. The patients/participants provided written informed consent to participate in this study. (16-18) In this study, WC, age, BMI, DM, ethnicity, gender, HBP, LDL, and smoking features were investigated. Some ethnicities in Iran are Gilak, Kurd, Turkmen, Azeri, Lur, Persian, Baloch, Qashqai, Arab, and Bakhtiari. Their population distribution is shown in Supplementary Figure 1.

## 3. Methods

### 3.1. Data augmentation

In this study, the data are unbalanced. The challenge of working with this type of data is that most classification algorithms ignore the minority class and consequently perform poorly. In contrast, the performance of the minority class is commonly more important. In these conditions, data augmentation can be used. This technique increases the training set artificially by creating modified copies of data using existing data in the dataset. This new sample generation may happen by making minor changes to the existing data or using other algorithms, such as deep learning, to generate new samples. In this study, the Synthetic Minority Over-sampling Technique or SMOTE upsampling (19) is used for data augmentation.

### 3.2. Feature weighting algorithms

Feature weighting algorithms help us to figure out which attribute(s) (feature(s) or variable(s)) are important to predict a given label (target) attribute. It is a type of feature engineering. Feature weighting methods are classified according to the used techniques for feature weighting. Statistical measures are one of the most common methods used to calculate the importance of attributes. Often used statistical measures are the Gini Index (20) and Information gain (21). On the other hand, the importance of attributes can be calculated by some models, such as rule induction and decision trees. In this study, some of the feature weighting algorithms such as Weight by Information Gain, Weight by Gini Index (20), Weight by Rule (22), Weight by Uncertainty (23), Weight by Relief (24), and Weight by Chi-Squared Statistics (25) were used.

### 3.3. Classification algorithms

A classification algorithm is a type of machine learning algorithm used to predict or identify the category or class of an input data point. In other words, it's an algorithm that uses pre-classified examples to develop a model to classify new instances. The goal of using these algorithms is to build a model that can be used to predict the class of an instance, given the instance's features. Some of the more common types of classification algorithms used in this study include Rule induction (26), Deep Learning algorithm (27), Gradient-boosted trees (28), Generalized Linear Models (29), Decision Tree (30), and Random Forest (31).

### 3.4. Proposed method

This study is divided into two parts. In the first part, using feature weighting algorithms, the relevance of each feature in label feature prediction is calculated. This is done for six different feature weighting algorithms, and finally, the average rank of each feature is computed using the rank of each feature in these different algorithms. In the second part, the data is divided using 10-fold cross-validation. Then, some of the most important classification algorithms were trained while ethnicity was/was not used as an input feature. The model created by each of these algorithms has been tested, and the performance of these algorithms is compared to determine not only the effect of using ethnicity in the performance of algorithms but also to find the best algorithms. Supplementary Figure 2 illustrates the steps of this study.

## 4. Experimental results

### 4.1. Weight of Attributes

In this section, at first, the results of different feature weighting algorithms, i.e., weight by information gain, weight by Gini Index, weight by Rule, weight by uncertainty, weight by uncertainty, weight by relief, and weight by Chi-Squared Statistics, are shown in Table 1. In this table, the corresponding row of ethnicity was bolded. The higher the weight of an attribute, the more relevant it is considered and the better the rank of it. According to this algorithm, the ranks and the weights of attributes are shown in Table 1. The average rank of each feature and its overall rank are shown in the two last columns of this table. Accordingly, gender, age, and ethnicity have better ranks with respect to other investigated features.

*Table 1. The rank and weight of each attribute are calculated by different feature weighting algorithm*

| Weight by<br>Attribute name | Information Gain | | Gini Index | | Rule | | Uncertainty | | Relief | | Chi-Squared Statistics | | Mean of Ranks | Overall Rank |
|---|---|---|---|---|---|---|---|---|---|---|---|---|---|---|
| | Rank | Weight | Rank | Weight | Rank | Weight | Rank | Weight | Rank | Weight | Rank | Weight | | |
| gender | 1.00 | 0.05733 | 1 | 0.03632 | 2 | 0.6477 | 1 | 0.05924 | 1 | 1.31518 | 1 | 259.3119 | 1.17 | 1 |
| age | 2.00 | 0.02003 | 2 | 0.01269 | 1 | 0.654 | 3 | 0.01518 | 9 | 0.00264 | 2 | 120.5562 | 3.17 | 2 |
| WC | 4.00 | 0.01753 | 3 | 0.01128 | 9 | 0.6407 | 2 | 0.01843 | 2 | 0.49711 | 3 | 80.56527 | 3.83 | 4 |
| **ethnicity** | **3.00** | **0.01826** | **4** | **0.01099** | **3** | **0.6425** | **5** | **0.0115** | **3** | **0.27172** | **4** | **79.42729** | **3.67** | **3** |
| smoking | 5.00 | 0.01258 | 5 | 0.00764 | 7 | 0.6407 | 4 | 0.01499 | 7 | 0.03802 | 5 | 54.58233 | 5.5 | 5 |
| DM | 6.00 | 0.01047 | 6 | 0.0065 | 5 | 0.6407 | 6 | 0.0115 | 8 | 0.00791 | 6 | 46.46549 | 6.17 | 6 |
| BMI | 7.00 | 0.00343 | 7 | 0.00221 | 8 | 0.6407 | 7 | 0.00372 | 6 | 0.07301 | 7 | 15.79585 | 7 | 8 |
| HBP | 8.00 | 0.00092 | 8 | 0.00058 | 4 | 0.6407 | 8 | 0.00095 | 4 | 0.13964 | 8 | 4.197016 | 6.67 | 7 |
| LDL | 9.00 | 0.00005 | 9 | 0.00003 | 6 | 0.6407 | 9 | 0.00005 | 5 | 0.10739 | 9 | 0.251706 | 7.83 | 9 |

### 4.2. Results of Classification Algorithms

This section shows the results of the classification algorithms used in this study. At first, ethnicity was not considered as an input variable. Then, it was added to input variables. As shown in Table 2 compared to Table 3, when ethnicity is used as an input variable, the average accuracy, precision, recall, and AUC are increased by 3.95%, 3.68%, 4.44%, and 0.04, respectively.

*Table 2. Classification results without ethnicity*

| Classification Algorithms / Criterion | Rule Induction | Deep Learning | Generalized Linear Model | Gradient Boosted Tree | Decision Tree | Random Forest | Average |
|---|---|---|---|---|---|---|---|
| accuracy | 72.92 ± 2.25 | 71.25 ± 1.05 | **73.35 ± 3.26** | 72.57 ± 2.37 | 71.54 ± 2.25 | 71.56 ± 2.38 | 72.20 ± 2.26 |
| precision | 75.63 ± 2.36 | 72.13 ± 1.79 | **75.26 ± 2.85** | 73.62 ± 1.75 | 73.26 ± 2.59 | 74.28 ± 1.25 | 74.03 ± 2.10 |
| recall | 83.55 ± 2.15 | **90.02 ± 3.24** | 86.84 ± 2.67 | 89.82 ± 3.64 | 86.85 ± 2.98 | 85.24 ± 2.33 | 87.05 ± 2.84 |
| AUC | 0.73 ± 0.03 | 0.75 ± 0.02 | **0.76 ± 0.03** | 0.75 ± 0.02 | 0.67 ± 0.03 | 0.74 ± 0.02 | 0.73 ± 0.03 |

*Table 3. Classification results with ethnicity*

| Classification Algorithms / Criterion | Rule Induction | Deep Learning | Generalized Linear Model | Gradient Boosted Tree | Decision Tree | Random Forest | Average |
|---|---|---|---|---|---|---|---|
| accuracy | 77.65 ± 1.22 | 75.54 ± 2.58 | **79.65 ± 1.77** | 75.24 ± 1.68 | 74.56 ± 1.79 | 74.24 ± 1.28 | 76.15± 1.72 |
| precision | 79.26 ± 3.24 | 76.85 ± 1.11 | **79.42 ± 1.45** | 77.26 ± 2.98 | 76.25 ± 2.24 | 77.23 ± 2.93 | 77.71 ± 2.32 |
| recall | 92.38 ± 3.28 | **94.92 ± 2.78** | 90.11 ± 3.55 | 93.85 ± 1.95 | 89.64 ± 1.19 | 88.05 ± 2.25 | 91.49 ± 2.50 |
| AUC | 0.77 ± 0.02 | 0.79 ± 0.03 | **0.81 ± 0.01** | 0.80 ± 0.03 | 0.69 ± 0.02 | 0.79 ± 0.02 | 0.77 ± 0.02 |

### 4.3. Feature rank analyses

In this section, the rank of the top five features in each ethnicity is shown in Figure 1. This ranking is based on the average ranking of the features weighting algorithms introduced in Section 3.2. For example, in the Azeri ethnicity, the top five features are age, gender, smoking, DM, and BMI. In ethnicities Azeri, Gilak, Kurd, Lur, and Arab, the age has the highest rank while in Bakhtiari, Qashqai, and all cases together, it has the second rank. Gender is another important feature. It has the first rank in Persian, Qashqai, and all the ethnicities together while in Kurd, Lur, Azeri, and Arab, it has the second rank. In the lower part of Figure 1, this ranking is done according to all the investigated cases as shown in Table 1.

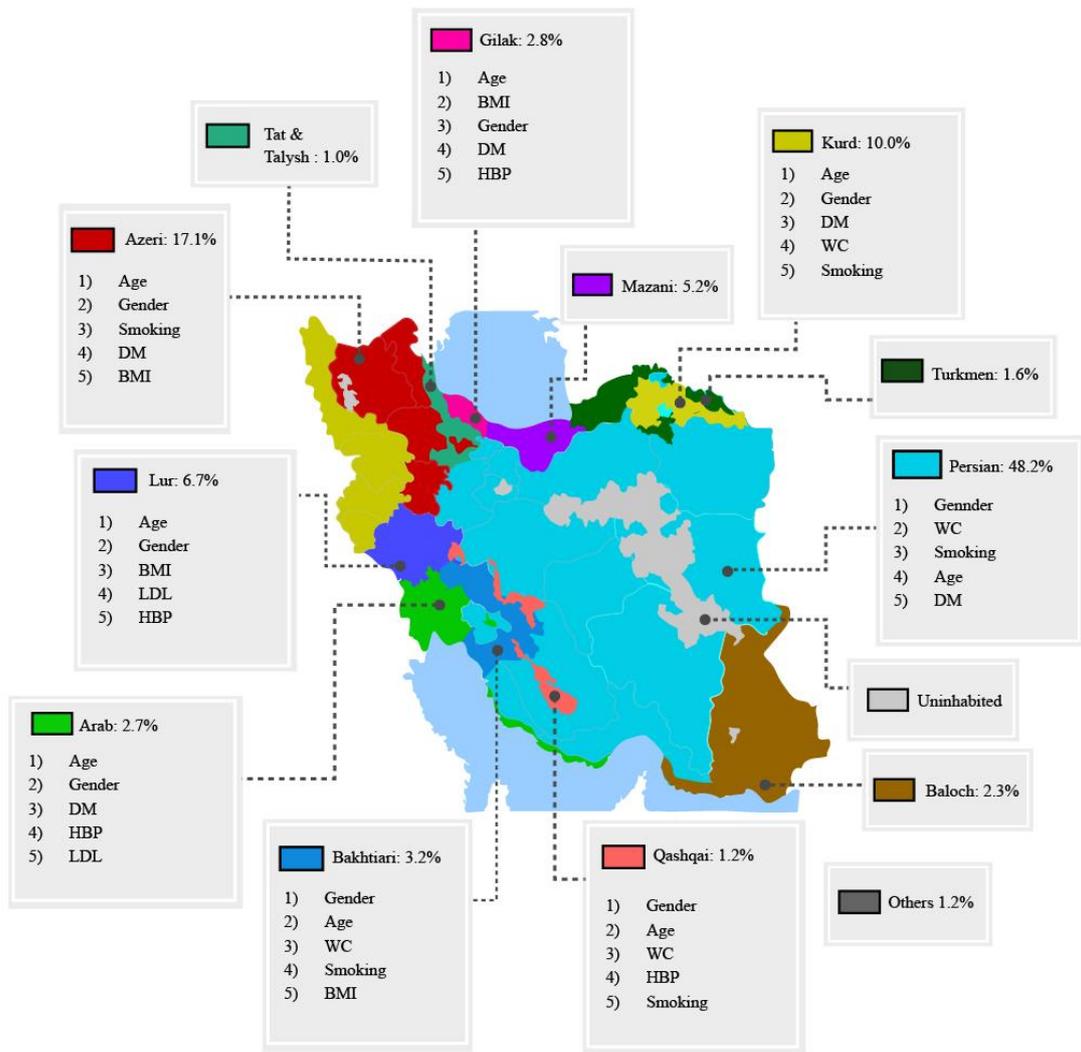

*Figure 1. The rank of the top features*

### 4.4. Analysis of the best classification algorithms

This section specifies the best classification algorithms based on each ethnicity. Their performance based on four factors: accuracy, precision, recall, and AUC are reported and shown in Figure 2. The best classification algorithm for all ethnicities is shown in the lower part of this figure. Accordingly, the algorithm Generalized Linear Model had the best efficiency in four ethnicities, Lur, Azeri, Gilak, and Persian, and in all the ethnicities together. In contrast, in the ethnicities Bakhtiari, Qashqai, and Kurd, the algorithm Gradient Boosted Trees, and in the ethnicity Arab, the algorithm Random Forest had a better efficiency. In addition, according to the accuracy rate, Kurd and Lur have the lowest and highest performances, respectively.

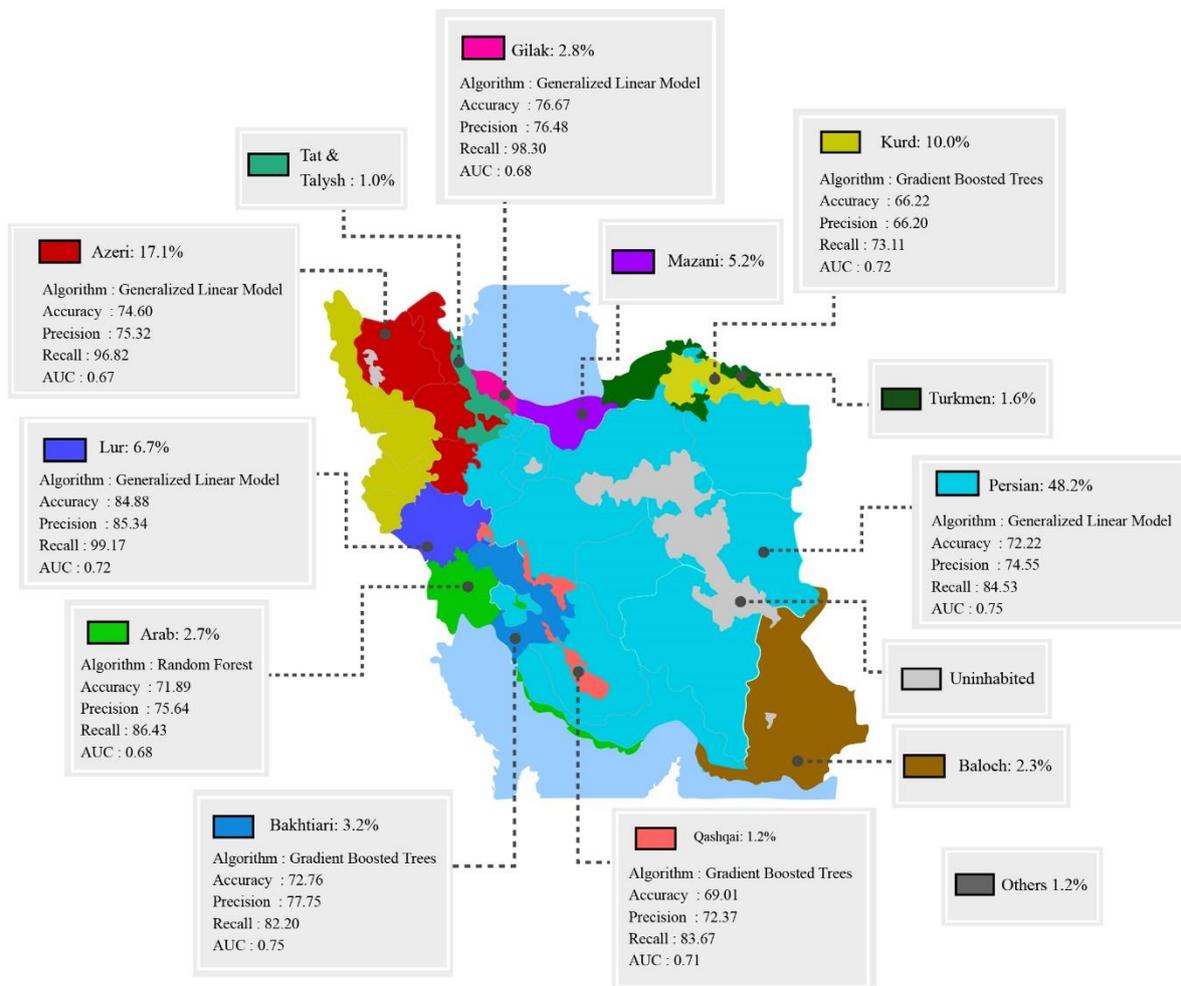

*Figure 2. The best classification algorithms based on each ethnicity*

## 5. Discussion

The study highlights the significance of ethnicity in predicting PCAD. It found that ethnicity ranks as the third most important risk factor among others like age and gender, emphasizing the need for tailored diagnostic approaches that consider ethnic backgrounds. Traditional methods for diagnosing CAD often overlook the nuances of different ethnicities, which can lead to misdiagnosis or ineffective treatment. This research aims to bridge that gap by utilizing machine learning algorithms incorporating ethnicity as a key feature [2]. The findings suggest that machine learning models can enhance predictive accuracy for PCAD when ethnicity is included as an input feature. This could lead to better risk stratification and more personalized healthcare interventions.

Additionally, the study's approach of comparing model performance with and without ethnicity provides valuable insights into how ethnic considerations can improve diagnostic outcomes. This could pave the way for future research to explore the impact of ethnicity across a broader range of cardiovascular diseases.

.

Overall, the research underscores the importance of integrating ethnic diversity into machine learning models for cardiovascular risk prediction, which could ultimately improve patient outcomes and healthcare equity.

# 6. Conclusion

PCAD, a type of heart disease, is primarily caused by the build-up of plaque in the walls of the coronary arteries. These arteries supply oxygen-rich blood to the heart muscle. It refers to the early onset of CAD, usually diagnosed before 55 years of age in men and 65 years of age in women. Risk factors for PCAD include WC, age, BMI, DM, gender, HBP, LDL, and smoking. In this study, the effect of ethnicity is investigated for PCAD diagnoses. According to the results, it ranks third among the investigated risk factors. Some classification algorithms were also used with/without ethnicity as one of the input features. The best way to continue this study is to investigate the effect of ethnicity on the diagnosis of PCAD in larger and more diverse ethnic populations.

**Acknowledgment:** The authors greatly appreciate the help of all staff in different study centers with their assistance in data collection and conducting intervention activities.

**Financial support:** This study was funded by the Research and Technology Department, Iran Ministry of Health and Medical Education, and Iranian Network of Cardiovascular Research.(#96110)

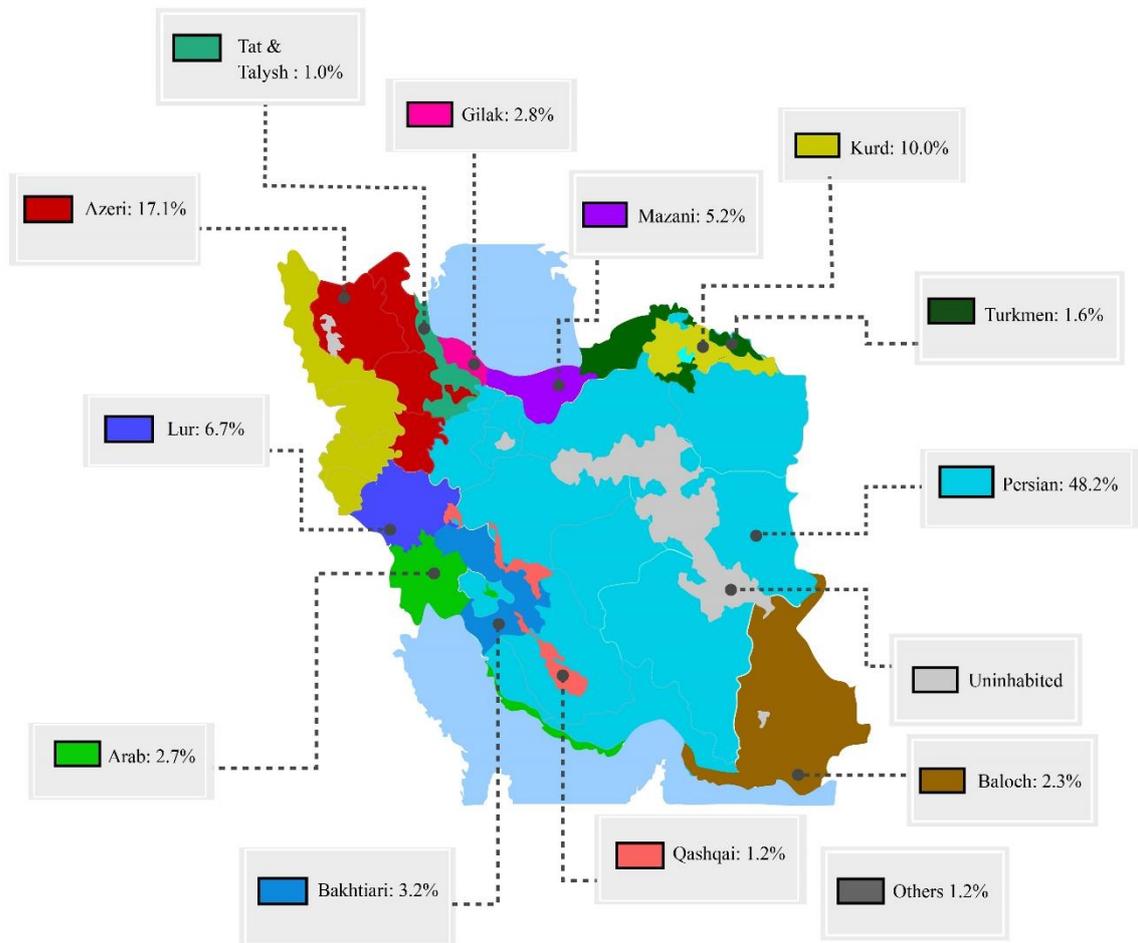

*Supplementary Figure 1. Ethnicity distribution in Iran*

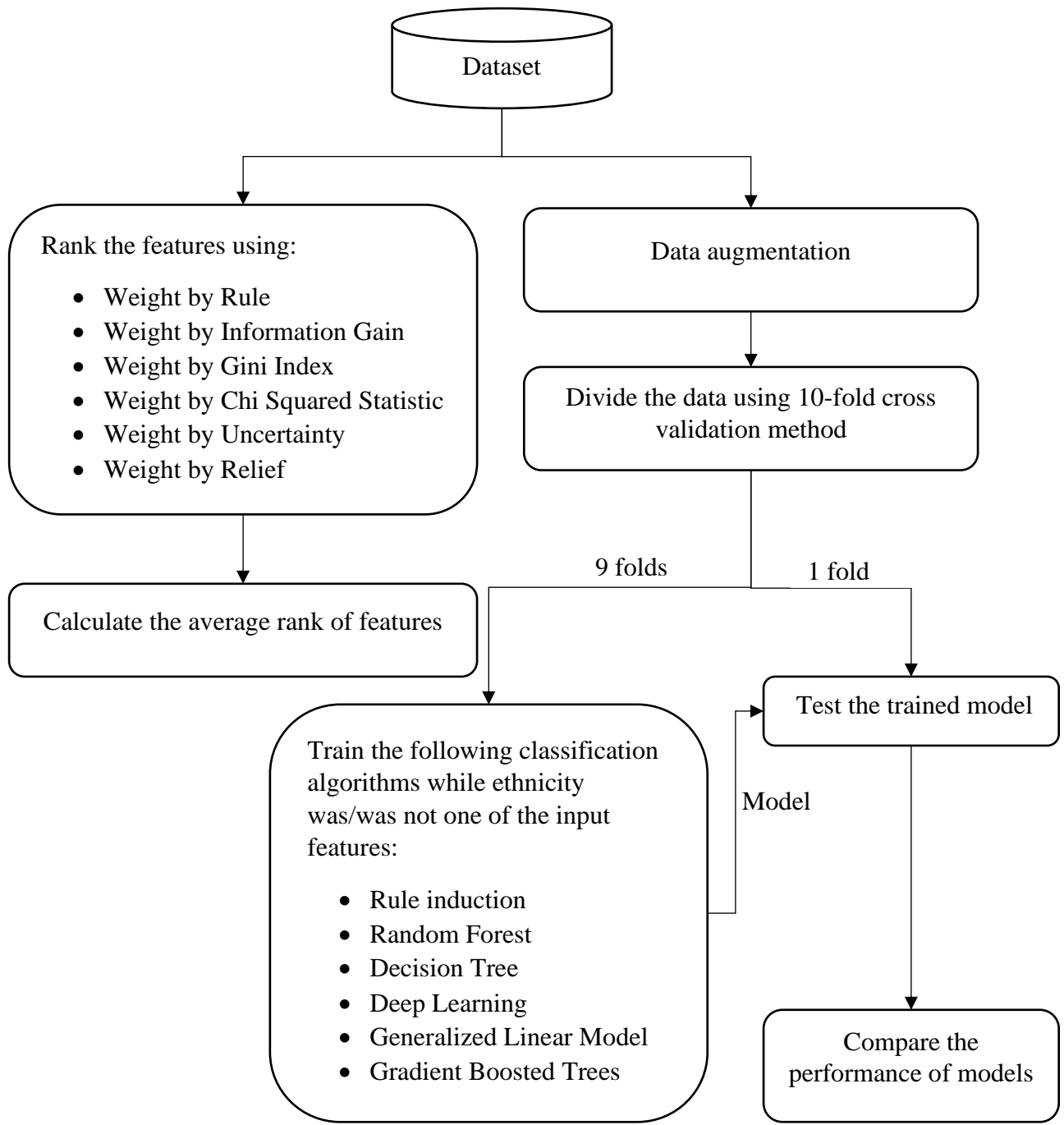

*Supplementary Figure 2. Flowchart of the proposed method*

Prevalence of the ethnic groups in Iran and IPAD study

| Ethnicity | Iran (%) | IPAD Study (%) |
|---|---|---|
| Fars (Persian) | 51 | 50 |
| Turk (Tork or Azari or Azeri) | 12 | 12.75 |
| Kurd (Kord) | 7-10 | 10 |
| Gilak | 3-6 | 6 |
| Lor | 3 | 3.5 |
| Arab | 3 | 3.5 |
| Bkhtiari | 1-2 | 3.5 |
| Qashqaei (Qashghai) | 1-2 | 3.5 |
| Balouch (Balouchi or Balooch) | 2 | 3.5 |

The total sample was divided proportionally according to the distribution of each ethnic group. Calculated sample for each ethnicity is as follows: 2000 Fars, 510 Azari, 400 Kurd, 250 Gilak, 140 Arab, 140 Lor, 140 Bakhtiari, 140 Qashghaei, and 140 Balouch patients

| Fars (n=1833) | Azari (n=303) | Kurd (n= 371) | Lor (n=180) | Bakhtiari (n= 209) | Qashqaei (n=133) | Gilak (n= 240) | Arab (n=98) | Balouch (n= 5) |
|---|---|---|---|---|---|---|---|---|